\documentclass[letterpaper]{article} % DO NOT CHANGE THIS
\usepackage{aaai2026}  % DO NOT CHANGE THIS
\usepackage{times}  % DO NOT CHANGE THIS
\usepackage{helvet}  % DO NOT CHANGE THIS
\usepackage{courier}  % DO NOT CHANGE THIS
\usepackage[hyphens]{url}  % DO NOT CHANGE THIS
\usepackage{graphicx} % DO NOT CHANGE THIS
\usepackage{natbib}  % DO NOT CHANGE THIS AND DO NOT ADD ANY OPTIONS TO IT
\usepackage{caption} % DO NOT CHANGE THIS AND DO NOT ADD ANY OPTIONS TO IT
\usepackage{algorithm}
\usepackage{algorithmic}

\usepackage{graphicx}
\usepackage{amsmath}
\usepackage{xcolor}
\usepackage{subcaption}
\usepackage{bbding}
\usepackage{multirow}
\usepackage{amssymb}
\makeatletter

\newcommand{\Rmnum}[1]{\expandafter\@slowromancap\romannumeral #1@}
\makeatother
\usepackage{pdfpages} 

\usepackage{newfloat}
\usepackage{listings}
\DeclareCaptionStyle{ruled}{labelfont=normalfont,labelsep=colon,strut=off} % DO NOT CHANGE THIS
\floatstyle{ruled}
\newfloat{listing}{tb}{lst}{}
\floatname{listing}{Listing}
\nocopyright
\title{Spatio-Temporal Distortion Aware Omnidirectional Video Super-Resolution}
\author{
    Hongyu An$^{1}$,
    Xinfeng Zhang$^1$\thanks{Corresponding author.},
    Shijie Zhao$^2$,
    Li Zhang$^2$,
    Ruiqin Xiong$^3$
}
\affiliations{
    \textsuperscript{\rm 1}University of Chinese Academy of Sciences,
    \textsuperscript{\rm 1}Bytedance Inc,
    \textsuperscript{\rm 1}Peking University\\
    anhongyu22@mails.ucas.ac.cn, 
    xfzhang@ucas.ac.cn, \\
    \{zhaoshijie.0526, lizhang.idm\}@bytedance.com,
    rqxiong@pku.edu.cn
}

\usepackage{bibentry}
\begin{document}

\maketitle

\begin{abstract}
Omnidirectional videos (ODVs) provide an immersive visual experience by capturing the $360^{\circ}$ scene. With the rapid advancements in virtual/augmented reality, metaverse, and generative artificial intelligence, the demand for high-quality ODVs is surging. However, ODVs often suffer from low resolution due to their wide field of view and limitations in capturing devices and transmission bandwidth. Although video super-resolution (SR) is a capable video quality enhancement technique, the performance ceiling and practical generalization of existing methods are limited when applied to ODVs due to their unique attributes. To alleviate spatial projection distortions and temporal flickering of ODVs, we propose a Spatio-Temporal Distortion Aware Network (STDAN) with joint spatio-temporal alignment and reconstruction. Specifically, we incorporate a spatio-temporal continuous alignment (STCA) to mitigate discrete geometric artifacts in parallel with temporal alignment. Subsequently, we introduce an interlaced multi-frame reconstruction (IMFR) to enhance temporal consistency. Furthermore, we employ latitude-saliency adaptive (LSA) weights to focus on regions with higher texture complexity and human-watching interest. By exploring a spatio-temporal jointly framework and real-world viewing strategies, STDAN effectively reinforces spatio-temporal coherence on a novel \emph{ODV-SR} dataset and ensures affordable computational costs. Extensive experimental results demonstrate that STDAN outperforms state-of-the-art methods in improving visual fidelity and dynamic smoothness of ODVs.
\end{abstract}


\section{Introduction}
Omnidirectional videos (ODVs), also known as 360$^{\circ}$ or panoramic videos, cover a full 360$^{\circ}$ field of view by stitching multiple images onto a spherical surface. This expansive vision enables engaging immersive interaction, making ODVs valuable in various fields, like entertainment, creativity, advertising, intelligent driving, and video conferencing. To achieve realistic visuals, ODVs ideally require high resolutions such as 4K, 8K, or higher \cite{resolution}. Nevertheless, the trade-off between display quality and transmission costs is challenging. Video super-resolution (VSR) offers a solution by reconstructing high-resolution (HR) frames from low-resolution (LR) ones. Compared with general videos, ODVs necessitate an additional projection step from sphere to plane for subsequent processing, which causes uneven stretching and broken boundaries. These issues render traditional VSR models ineffective for ODVs.

Practical ODV projection methods from 3D to 2D include Equirectangular Projection (ERP), Cubemap Projection (CMP), Icosahedral Projection (ISP), and Equi-Angular Cubemap Projection (EAC), {\em etc.} Among these, ERP is the most widely used due to its low computation complexity and extensive adaptability. As illustrated in Fig. \ref{fig.1}(a)\textcolor{red}{\footnote{Detailed derivative processes can be found in Appendix 1.}}, spherical coordinates are defined as ${(\rho, \theta, \phi)}$, where ${\theta \in (0, 2\pi)}$ and ${\phi \in (0, \pi)}$ represent longitude and latitude, respectively. The planar horizontal and vertical coordinates are defined as ${(u, v)}$. However, ERP notably introduces latitude-related distortions, particularly severe in polar areas. Simultaneously, ERP creates discontinuities at rectangular edges, disrupting the continuity of adjacent pixels on the sphere.

\begin{figure}[!t]
\centering
\subfloat[Equirectangular Projection (ERP).]
{\includegraphics[width=0.95\columnwidth]{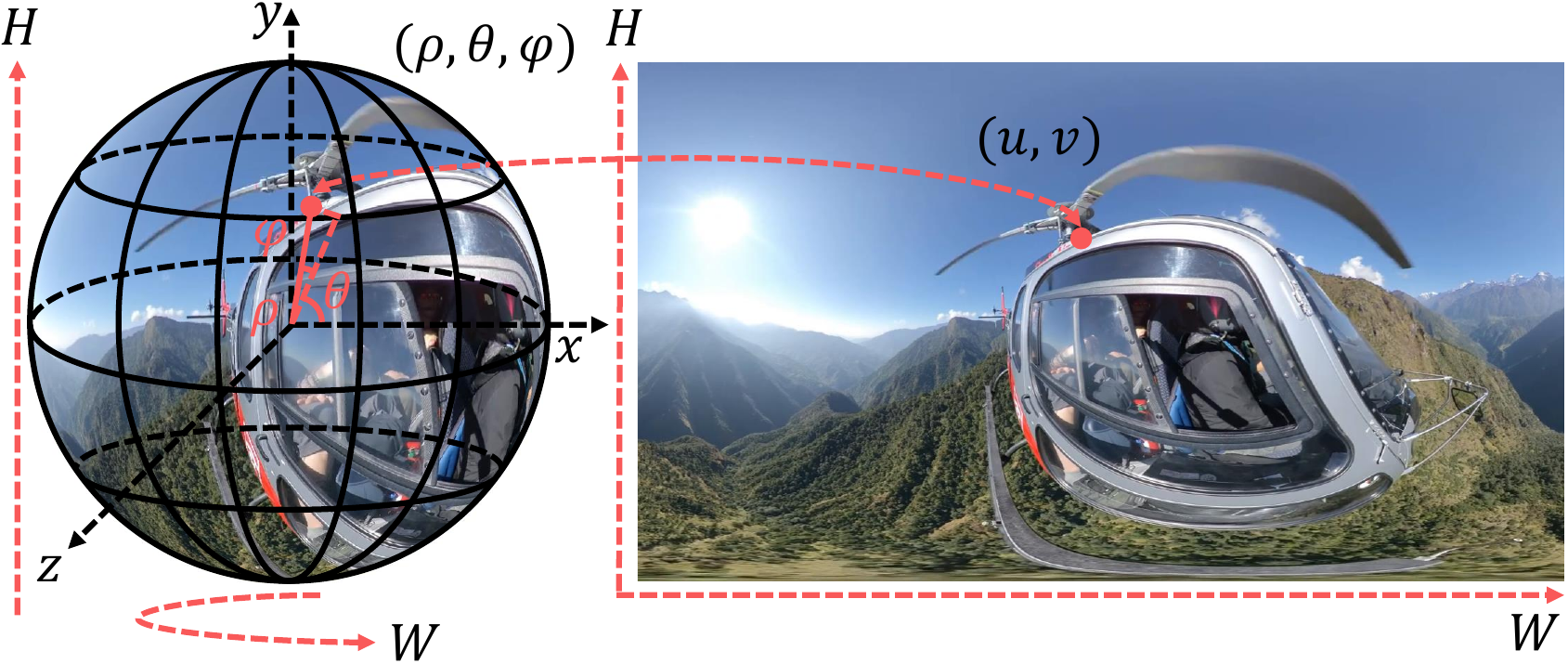}} \\
\vspace{3pt}
\subfloat[ERP viewpoint conversion.]{\includegraphics[width=0.95\columnwidth]{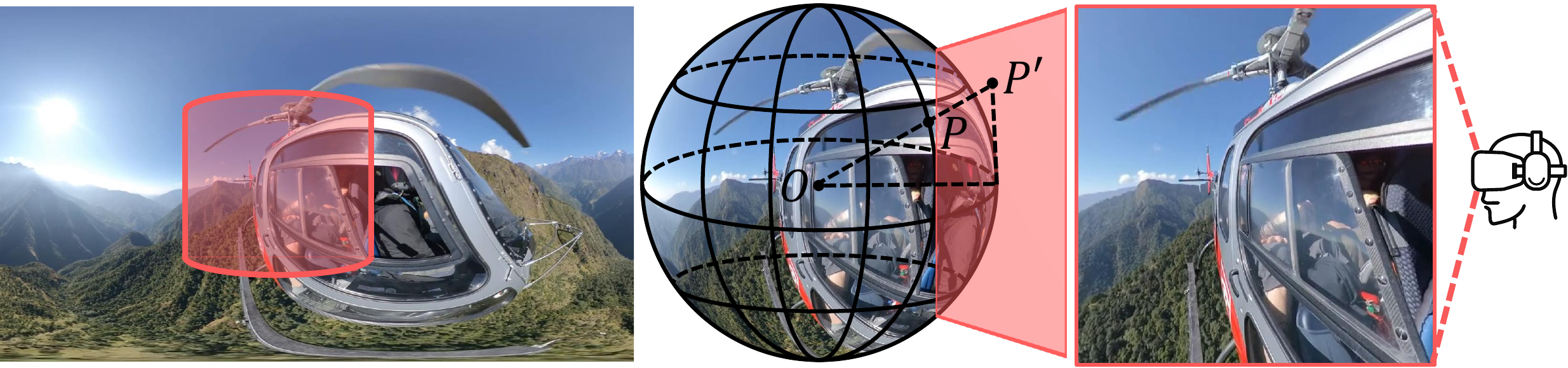}}
\vspace{-7pt}
\caption{Geometric transformations: (a) ERP. (b) A practically observed ERP viewpoint. To align with viewing preferences, attractive viewpoints should be emphasized.}
\vspace{-14pt}
\label{fig.1}
\end{figure}

Deep learning-based VSR models have replaced traditional interpolation methods, achieving remarkable success. Unfortunately, these techniques fall short when applied to ODVs with spatio-temporal distortions. As for omnidirectional image (ODI) super-resolution (SR), existing methods focused on reducing ERP distortions through latitude correlation (Ozcinar et al. 2019; Deng et al. 2021; An and Zhang 2023). OSRT \cite{OSRT} designed feature-level offset and alignment. FATO \cite{FATO} explored frequency distribution. Research on ODV-SR remains limited, with current works \cite{SMFN, S3PO, VertexShuffle} concentrating on latitude-related distortions without considering spatial discontinuity and temporal inconsistency. Moreover, the aforementioned models overlook the human visual pattern. As demonstrated in Fig. \ref{fig.1}(b), the highlighted region stands for the real viewpoint of interest.

%Meanwhile, the emerging Transformer-based architecture results in an inevitable increase in computational complexity, restricting the applicability in scenarios demanding extremely HR ODVs.

\begin{figure}[!t]
\centering
\includegraphics[width=0.95\columnwidth]{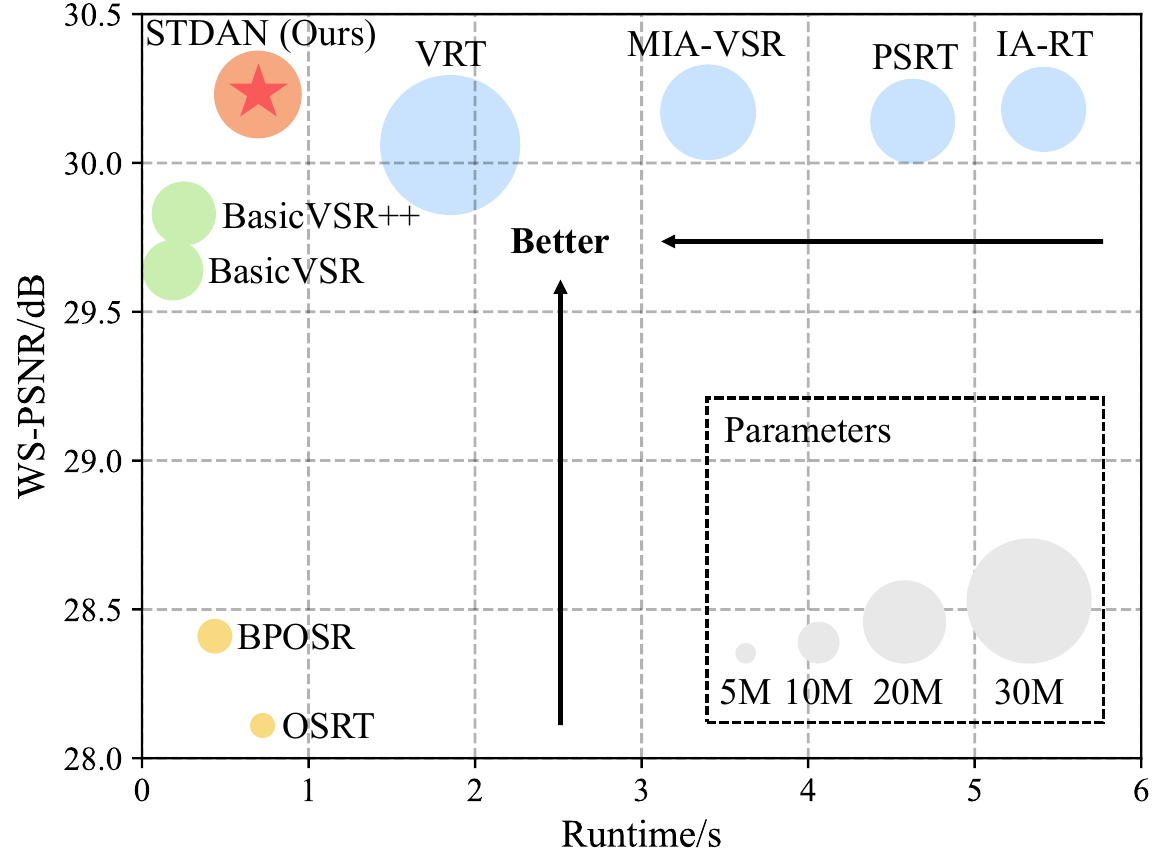}
\vspace{-6pt}
\caption{WS-PSNR(dB) and Runtime(s) comparison. Our STDAN outperforms other methods and strikes a balance between performance and computational efficiency.
}
\label{fig.2}
\vspace{-12pt}
\end{figure}

This paper proposes a Spatio-Temporal Distortion Aware Network (STDAN) to address the unique challenges of ODV-SR. While general VSR methods exploit inter-frame similarity through temporal alignment, extending beyond spatial feature extraction and up-sampling reconstruction, they often fail to handle the complex spatio-temporal degradations in ODVs. STDAN optimizes single-dimensional spatial or temporal modules and integrates joint spatio-temporal alignment and reconstruction. Specifically, it leverages spatial alignment with omni-positional encoding priors to modulate projection stretching and discontinuous boundaries, and introduces an interlaced multi-frame reconstruction module to maintain temporal stability. Moreover, latitude-saliency adaptive weights are computed within the loss function to concentrate on high-heat regions. We also present a new ODV-SR dataset with diverse scenarios. Both objective metrics and subjective evaluations demonstrate the superior performance of STDAN over existing methods. The main contributions can be summarized as follows: 
\begin{itemize}
\item[1)] We propose STDAN to fully exploit the spatio-temporal properties of ODVs. The spatio-temporal continuous alignment (STCA) incorporates positional information to mitigate projection distortions effectively.
\item[2)] An interlaced multi-frame reconstruction (IMFR) is designed based on a spatio-temporal sequence rescheduling mechanism, guaranteeing temporal consistency.
\item[3)] A latitude-saliency adaptive (LSA) loss is adopted to enhance regions with high texture complexity and watching interest, thereby enhancing perceptual quality.
\item[4)] We develop the first ODV-SR network tailored for practical applications, emphasizing computational efficiency and focusing on actual viewpoints, as shown in Fig. 2. Additionally, a variety of ODV-SR dataset is introduced to improve the generalization capability of STDAN.
\end{itemize}

\section{Related Work}

\subsection{Video Super-Resolution}
Traditional VSR networks are broadly divided into two categories: sliding-window and recurrent frameworks. The former \cite{VSRnet, VESPCN, EDVR, TDAN} leverages the temporal correlation between a reference frame and its neighboring frames within a short temporal window. The latter \cite{FSTRN, RBPN, BasicVSR, BasicVSR++} employs a recurrent mechanism to capture long-term relationships across frames. Benefiting from the long-term modeling capability, Transformer \cite{Transformer} has been integrated into VSR \cite{VRT, Patch, MIAVSR, IART, BT-STVSR}. However, the complex attention calculations in Transformer-based VSR approaches pose significant challenges for practical deployment, particularly when it comes to enhancing ODVs to at least 2K resolution.

\subsection{Omnidirectional Image Super-Resolution}
Intuitively, Ozcinar \textit{et al.} \cite{360-SS} designed a latitude-related loss to alleviate ERP distortions. LAU-Net \cite{LAU-Net} designed a framework with multiple latitude levels. SphereSR \cite{SphereSR} proposed an icosahedron-based feature extraction module. OSRT \cite{OSRT} presented a distortion-aware Transformer oriented to real-world fisheye down-sampling. FATO \cite{FATO} reduced ODI distortions by analyzing the frequency domain distribution. BPOSR \cite{BPOSR} applied a bi-projection network to facilitate the fusion between different projections. FAOR \cite{FAOR} adapted the implicit image function from the planar domain to the ERP domain by integrating spherical geometric priors.

\begin{figure*}[!t]
\centering
\includegraphics[width=2\columnwidth]{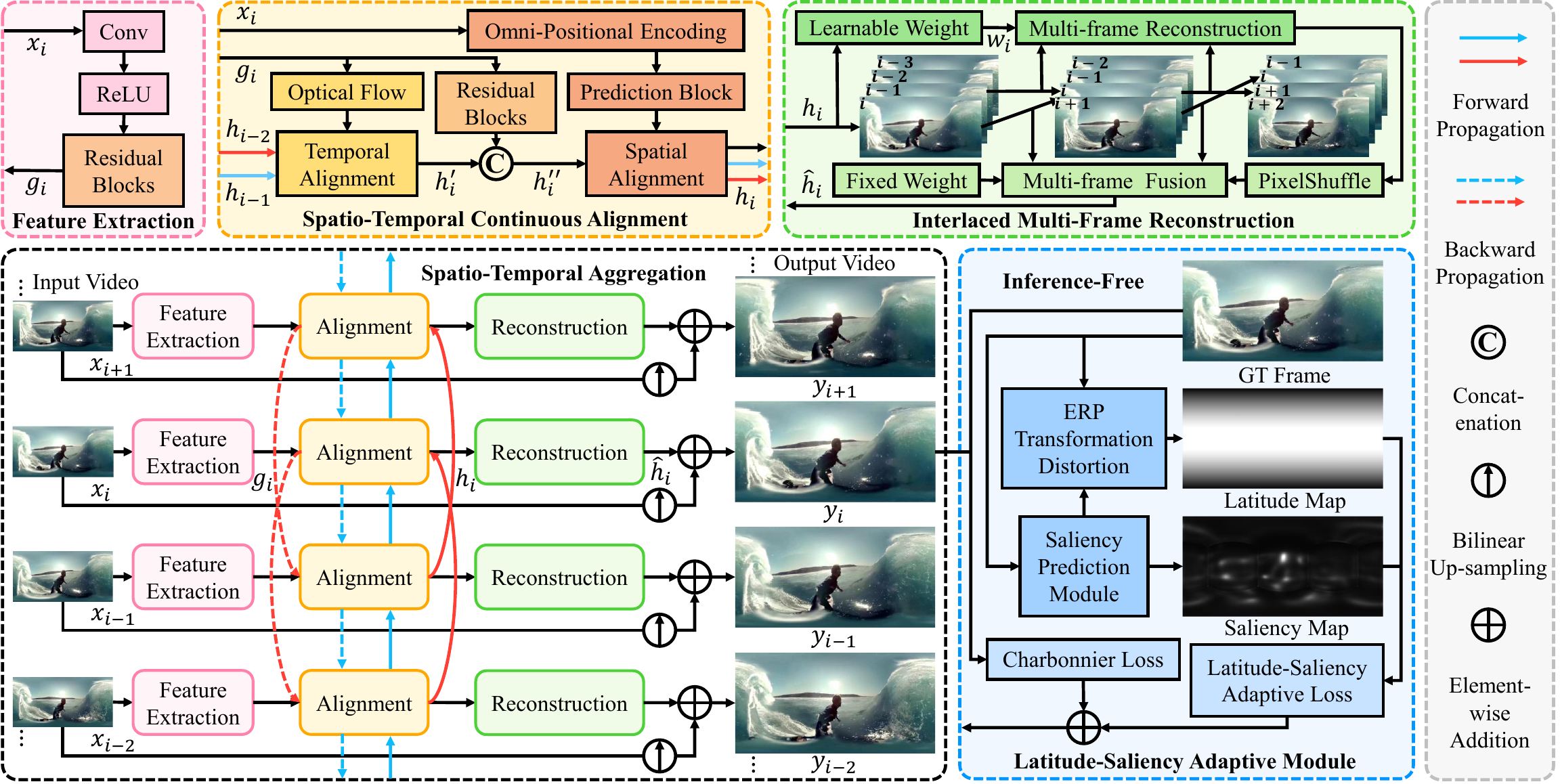}
\vspace{-4pt}
\caption{Overview of the proposed Spatio-Temporal Distortion Aware Network (STDAN).}
\label{fig_3}
\vspace{-8pt}
\end{figure*}

\subsection{Omnidirectional Video Super-Resolution}
Despite the impressive progress in VSR techniques, ODV-SR has not garnered the attention it deserves. SMFN \cite{SMFN} introduced a dual single-frame and multi-frame network. A weighted loss function is also provided to make the network pay more attention to equatorial regions. S3PO \cite{S3PO} adopted a spherical distortion feature extractor in an attention-recurrent framework, unbound from conventional VSR alignment. VertexShuffle \cite{VertexShuffle} utilized a focused icosahedral mesh to represent local sphere regions and further constructed matrices to rotate spherical content over the entire sphere. Although referenced ODI/ODV-SR models modulated latitude-related ERP distortions effectively, the spatial discontinuity inherent in projection and temporal inconsistency remain underexplored. Besides, the tendency of human visual attention in ODVs is hardly involved. In this paper, we propose an efficient spatio-temporal aggregation framework with watching inclination to enhance the immersive experience. 

\section{Methodology}

\subsection{Overview}
The overall framework of STDAN is depicted in Fig. \ref{fig_3}. It is built upon the bi-directional second-order grid propagation framework \cite{BasicVSR++} and comprises: spatio-temporal continuous alignment (STCA), interlaced multi-frame reconstruction (IMFR), and a latitude-saliency adaptive (LSA) module. Given a LR video sequence ${\{x_1, ..., x_i, ..., x_n\}}$, we first extract the feature ${g_i}$ from each frame ${x_i}$. This feature ${g_i}$ is then fed into STCA, where it undergoes spatio-temporal alignment to produce an enhanced feature ${h_{i}}$. The process is formulated as follows:
\begin{equation}
\label{equ_1}
h_{i} = \text{SA}(\text{C}(\text{TA}(h_{i-2}, h_{i-1}, g_{i}), \text{Res}(g_{i})),
\end{equation}
where TA($\cdot$) and SA($\cdot$) refer to temporal and spatial alignments, C($\cdot$) stands for concatenation along channel dimension, and Res($\cdot$) denotes residual blocks. Subsequently, the enhanced feature ${h_{i}}$ is processed by the IMFR module for interlaced spatio-temporal reconstruction, as described by:
\begin{equation}
\label{equ_2}
\widehat{h}_{i} = \text{MFF}(\text{PixelShuffle}(\text{MFR}(h_i, w_i)), w^{'}_{i}),
\end{equation}
where MFR($\cdot$) and MFF($\cdot$) indicate the multi-frame reconstruction and fusion processes. The parameter $w_{i}$ is a learnable weight that guides interaction across frames. After MFR, we employ PixelShuffle \cite{PixelShuffle} to up-sample reconstructed features and further smooth outputs using a fixed-weight $w^{'}_{i}$. The final HR frame ${y_i}$ is generated via a global residual connection. Notably, we estimate the LSA loss to emphasize practically appealing regions.

\subsection{Spatio-Temporal Continuous Alignment (STCA)}
The ERP inherently introduces distortions and discontinuities, which pose challenges for ODV-SR. To address these issues, we propose STCA that integrates omni-positional encoding (OPE) to achieve joint spatio-temporal alignment.

\begin{figure}[!t]
\centering
\subfloat[Temporal Alignment]{\includegraphics[width=0.65\columnwidth]{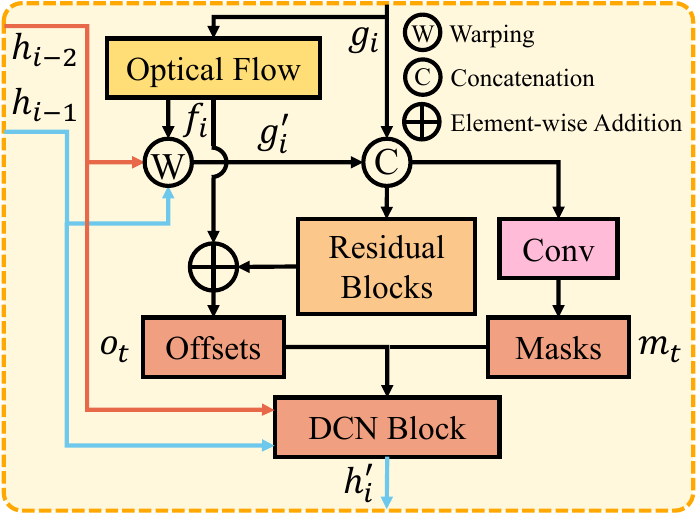}}
\hspace{2pt}
\vspace{-4pt}
\subfloat[Spatial Alignment]{\includegraphics[width=0.33\columnwidth]{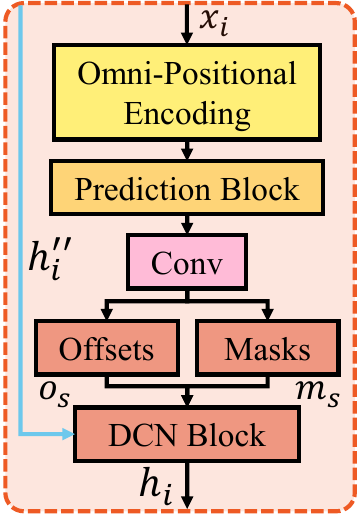}}
\vspace{-2pt}
\caption{DCN-based temporal correlation alignment (a) and spatial distortion alignment (b). Optical flows and omni-positional encoded offsets supply spatio-temporal priors.}
\label{fig_4}
\vspace{-13pt}
\end{figure}

\subsubsection{Dimension Expansion Alignment}
DCN-based \cite{DCN} temporal alignment has been widely adopted to compensate for motions. Following BasicVSR++ \cite{BasicVSR++}, we apply SpyNet \cite{SpyNet} to compute the optical flow $f_i$ as temporal guidance. Based on this optical flow, we calculate the offset $o_t$ and mask $m_t$ of the warped feature $g^{'}_{i}$, which serve as the initialization for temporal $\text{DCN}_\text{T}$. Fig. \ref{fig_4} (a) inspects the forward alignment, with the backward progress following a similar procedure. To tackle geometric distortions in ODVs, we extend the alignment range. Unlike previous feature-level warping \cite{OSRT}, we design a spatio-temporal joint framework with an ODV-oriented OPE to achieve multi-dimensional continuous modulation. As shown in Fig. \ref{fig_4} (b), we incorporate positional information and estimate distortions through a prediction block, which consists of a 1 ${\times}$ 1 convolution with two hidden layers. The proposed STCA is defined as follows:
\begin{equation}
o_t=\text{Res}(\text{C}({g}_{i}, {g}^{'}_{i})) + {f}_{i},m_t=\text{Conv}(\text{C}({g}_{i}, {g}^{'}_{i})),
\end{equation}
\begin{equation}
\scalebox{0.93}{$\displaystyle
{h}^{'}_{i} = \text{DCN}_{\text{T}}(\text{C}({h}_{i-1}, {h}_{i-2}), o_t, m_t),{h}^{''}_{i} = \text{C}({h}^{'}_{i}, \text{Res}({g}_{i})),
$}
\end{equation}
\begin{equation}
o_s=\text{Conv}(\text{P}(\text{OPE}({x}_{i}))),m_s=\text{Conv}(\text{P}(\text{OPE}({x}_{i}))),
\end{equation}
%\vspace{-3pt}
\begin{equation}
h_{i} = \text{DCN}_\text{S}({h}^{''}_{i}, o_s, m_s),
\end{equation}
where OPE($\cdot$) means omni-positional encoding, and P($\cdot$) represents prediction block. The prediction block allows positional priors to adapt dynamically to the content distribution, thereby enhancing the flexibility of spatial DCN.

\begin{figure}[!t]
\centering
\subfloat[ERP boundary and horizontal PE.]{\includegraphics[width=\columnwidth]{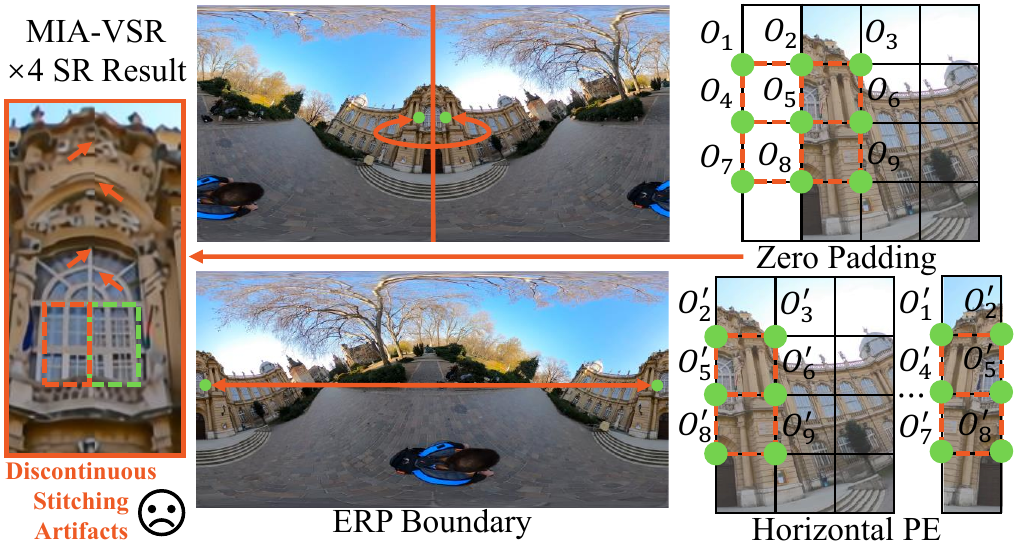}} \\
\vspace{2pt}
\subfloat[Sampling location of spatial DCN.]{\includegraphics[width=0.95\columnwidth]{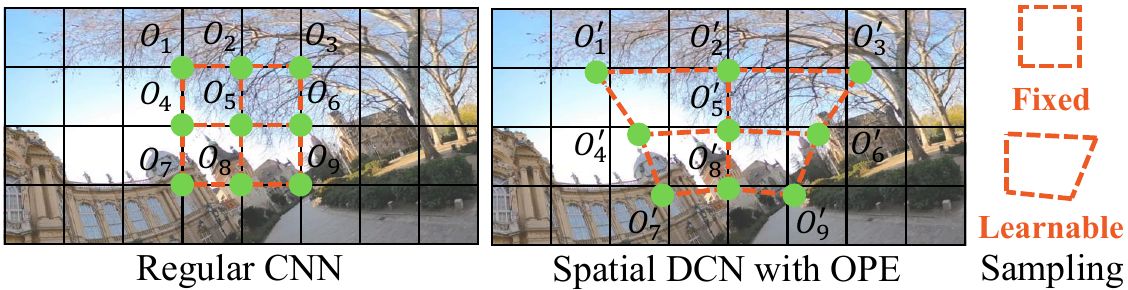}}
\vspace{-4pt}
\caption{Spatial alignment with OPE. Extra positional information enables continuous sampling.}
\label{fig_5}
\vspace{-10pt}
\end{figure}

\subsubsection{Omni-Positional Encoding}
In contrast to the complicated cylinder-style convolution proposed for ODI inpainting \cite{Cylin}, we develop a tight spatial alignment with OPE to exploit geometric priors. Specifically, we capture ERP positional cues to predict offsets for a learnable spatial DCN, achieving precise distortion modeling. For ODVs, ERP introduces quantifiable latitude-dependent distortions (Appendix 1), hence, we adopt cosine-based vertical PE at each latitude. Additionally, the ERP left-right boundary separates adjacent pixels and introduces stitching artifacts. As depicted in Fig. \ref{fig_5} (a), boundary windows exhibit severe mismatches, degrading visual quality. Since polar regions are free of discontinuities, we introduce a horizontal PE that mimics the unrolling of a cylinder:
\begin{equation}
[sin(1/10000^{2k/d}u),cos(1/10000^{2k/d}u),\dots],
\end{equation}
where $d$ indicates one-half of the encoding dimensions. This handcrafted encoding explicitly facilitates interaction across boundaries. Finally, we concatenate vertical and horizontal PE to obtain the complete OPE. After OPE-guided prediction and learnable spatial DCN, STCA adaptively reallocates sampling positions to modulate ERP distortions and restore spatial coherence, as illustrated in Fig. \ref{fig_5} (b).

\subsection{Interlaced Multi-Frame Reconstruction (IMFR)}
The memory constraint of HR ODVs limits the sequence length that can be processed. It is intuitive to divide the entire sequence into short subsequences and enhance them sequentially. However, truncated frames risk flickering. To improve visual consistency, we explore the relationship across restricted frames in subsequences and propose an interlaced multi-frame reconstruction (IMFR). In brief, IMFR reconstructs three temporally interlaced sequences and gets target frames through sliding interaction among corresponding frames in three sequences.

\begin{figure}[!t]
\centering
\includegraphics[width=\columnwidth]{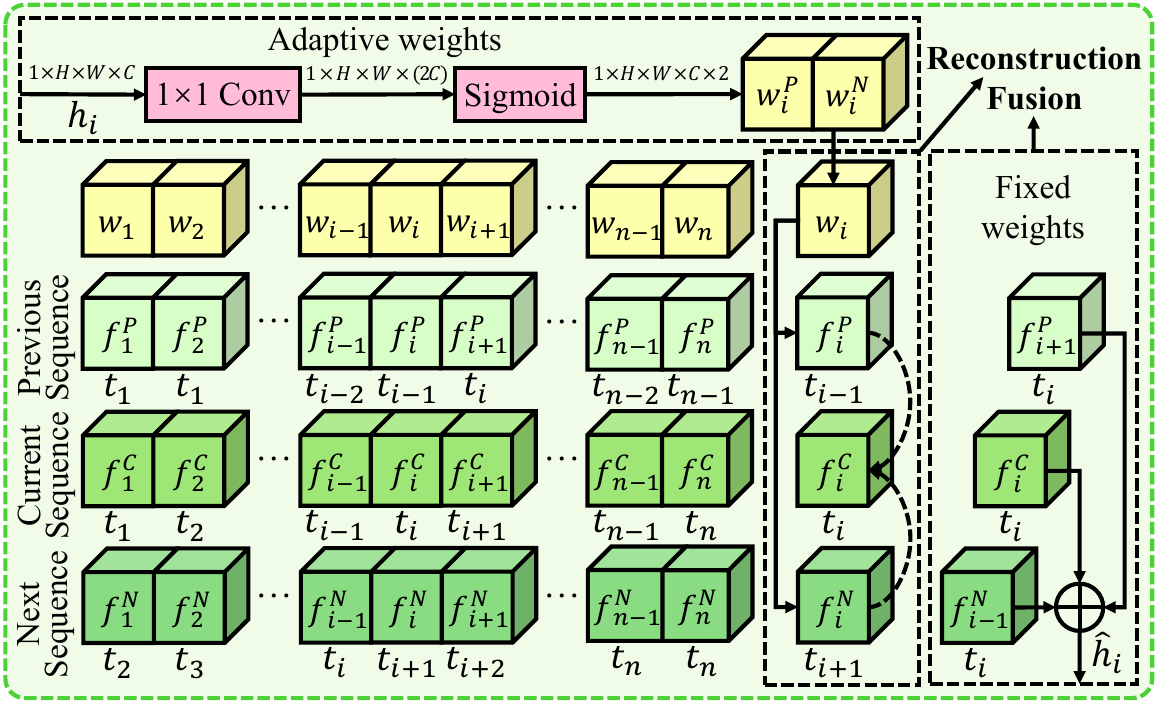}
\vspace{-15pt}
\caption{Interlaced multi-frame reconstruction. The target frame ${f^{C}_{i}}$ is reconstructed with weighted ${f^{P}_{i}}$ and ${f^{N}_{i}}$ and fused with ${f^{P}_{i+1}}$ and ${f^{N}_{i-1}}$ to ensure temporal consistency.}
\label{fig_6}
\vspace{-12pt}
\end{figure}

To leverage temporal correlation among limited frames, we triple the channel of intermediate features and split them along the temporal axis into three subsequences: previous ${f^{P}}$, current ${f^{C}}$, and next ${f^{N}}$. Each index $i$ thus corresponds to prior frame, target frame, and subsequent frame. To unify sequence lengths, we pad ${f^{P}}$ and ${f^{N}}$ by replicating their first and last frames. As demonstrated in Fig. \ref{fig_6}, the target frame ${f^{C}_{i}}$ is jointly reconstructed from its neighbors ${f^{P}_{i}}$ and ${f^{N}_{i}}$. The aligned feature ${h_{i}}$ is fed into a 1 ${\times}$ 1 convolution to predict adaptive weights ${\emph{w}^{P}_{i}}$ and ${\emph{w}^{N}_{i}}$ that modulate the contributions of ${f^{P}_{i}}$ and ${f^{N}_{i}}$. A Sigmoid followed by a scaling factor of 0.5 constrains the range of weights.

Thereafter, we up-sample three sequences and fuse the target frame ${f^{C}_{i}}$ with its temporal counterparts ${f^{P}_{i+1}}$ and ${f^{N}_{i-1}}$, enforcing a weighted smoothing across the three subsequences. Formally, the fused frame is computed via:
\begin{equation}
\label{equ_13}
f_{i} = \text{Norm}(\alpha_1 f^{P}_{i+1} + f^{C}_{i} + \beta_1 f^{N}_{i-1}),
\end{equation}
where Norm($\cdot$) indexes normalization, and weights $\alpha_1$ and $\beta_1$ are preset to 0.01. The IMFR propagates temporal context into every frame and smooths it, yielding a temporally coherent video sequence.

\begin{table*}[!t]
\centering
\footnotesize
\renewcommand\arraystretch{0.84}{
\setlength{\tabcolsep}{7.5pt}
\begin{tabular}{c|c|c|c|c|c}
\hline
Dataset&\multicolumn{5}{c}{\emph{ODV-SR} Dataset}\\
\hline
Method&PSNR (dB)$\uparrow$ / SSIM$\uparrow$&WS-PSNR (dB)$\uparrow$ / WS-SSIM$\uparrow$&$\text{E}_{warp}^*$$\downarrow$&VMAF$\uparrow$&Top-5 PSNR (dB)$\uparrow$ / SSIM$\uparrow$\\
\hline
Bicubic&28.01 / 0.8218&27.20 / 0.7933&\underline{8.25}&44.85&26.37 / 0.7562\\
%360-SS&28.56 / 0.8362&27.81 / 0.8114&10.17&63.70&27.51 / 0.7998\\
OSRT&27.98 / 0.8419&28.11 / 0.8314&9.84&70.00&28.01 / 0.8204\\
BPOSR&27.58 / 0.8024&28.41 / 0.8294&9.65&71.58&28.27 / 0.8148\\
EDVR&30.40 / 0.8765&29.61 / 0.8569&11.69&81.56&29.87 / 0.8580\\
BasicVSR&30.53 / 0.8785&29.64 / 0.8591&8.42&82.54&29.76 /  0.8581\\
BasicVSR++&30.72 / 0.8807&29.83 / 0.8621&8.44&83.39&30.10 / 0.8640\\
VRT&30.92 / 0.8832&30.06 / 0.8649&8.74&\underline{83.88}&30.44 / 0.8665\\
PSRT&31.01 / 0.8834&30.14 / 0.8653&8.68&83.77&\underline{31.53} / 0.8732\\
MIA-VSR&31.09 / 0.8843&30.17 / \underline{0.8660}&8.71&83.57&31.48 / 0.8729\\
IA-RT&\underline{31.11} / \underline{0.8845}&\underline{30.18} / 0.8659&8.69&83.75&31.47 / \underline{0.8734}\\
STDAN(Ours)&\textbf{31.13} / \textbf{0.8876}&\textbf{30.23} / \textbf{0.8698}&\textbf{7.60}&\textbf{85.13}&\textbf{30.61} / \textbf{0.8736}\\
\hline
Dataset&\multicolumn{2}{c|}{360VDS Dataset}&\multicolumn{3}{c}{MiG Dataset}\\
\hline
Method&PSNR(dB)$\uparrow$ / SSIM$\uparrow$&WS-PSNR(dB)$\uparrow$ / WS-SSIM$\uparrow$&\multicolumn{2}{c}{PSNR(dB)$\uparrow$ / SSIM}$\uparrow$&WS-PSNR(dB)$\uparrow$ / WS-SSIM$\uparrow$\\
\hline
Bicubic&25.11 / 0.7518&24.37 / 0.7164&\multicolumn{2}{c|}{28.97 / 0.8215}&28.78 / 0.7985\\
%360-SS&25.80 / 0.7791&25.04 / 0.7402&\multicolumn{2}{c|}{29.67 / 0.8359}&29.37 / 0.8143\\
OSRT&25.35 / 0.7926&25.00 / 0.7676&\multicolumn{2}{c|}{30.63 / 0.8618}&30.13 / 0.8472\\
EDVR&26.79 / 0.8252&25.99 / 0.7994&\multicolumn{2}{c|}{30.30 / 0.8534}&29.93 / 0.8340\\
%RBPN\cite{RBPN}&27.21 / 0.8328&26.32 / 0.8075&30.16 / 0.8446&29.76 / 0.8319\\
SFMN&-&-&\multicolumn{2}{c|}{30.56 / 0.8505}&30.13 / 0.8381\\
BasicVSR&27.10 / 0.8313&26.25 / 0.8057&\multicolumn{2}{c|}{30.71 / 0.8588}&30.10 / 0.8397\\
BasicVSR++&27.21 / 0.8358&26.36 / 0.8110&\multicolumn{2}{c|}{30.71 / 0.8578}&30.23 / 0.8387\\
S3PO&27.24 / 0.8225&26.31 / 0.8026&\multicolumn{2}{c|}{31.16 / 0.8565}&30.42 / 0.8453\\
VRT&27.23 / 0.8359&26.43 / 0.8124&\multicolumn{2}{c|}{30.85 / 0.8636}&\underline{30.45} / 0.8461\\
PSRT&27.30 / 0.8383 &26.48 / 0.8148&\multicolumn{2}{c|}{30.97 / 0.8625}&30.42 / 0.8443\\
MIA-VSR&27.28 / 0.8380&26.48 / 0.8146&\multicolumn{2}{c|}{31.03 / 0.8619}&30.38 / 0.8433\\
IA-RT&\underline{27.34} / \underline{0.8397}&\textbf{26.53} / \underline{0.8164}&\multicolumn{2}{c|}{\textbf{31.24} / \underline{0.8655}}&\textbf{30.50} / \underline{0.8472}\\
STDAN(Ours)&\textbf{27.38} / \textbf{0.8408}&\underline{26.51} / \textbf{0.8165}&\multicolumn{2}{c|}{\underline{31.21} / \textbf{0.8669}}&\textbf{30.50} / \textbf{0.8488}\\
\hline
\end{tabular}}
\vspace{-2pt}
\caption{Quantitative $\times$4 ODV-SR comparison on \emph{ODV-SR}, 360VDS, and MiG Panorama Video datasets. \textbf{Bold} and \underline{underlined} values indicate the best and second-best results.}\label{tab_ODV_SR}
\vspace{-10pt}
\end{table*}

\begin{figure}[!t]
\centering
\includegraphics[width=\columnwidth]{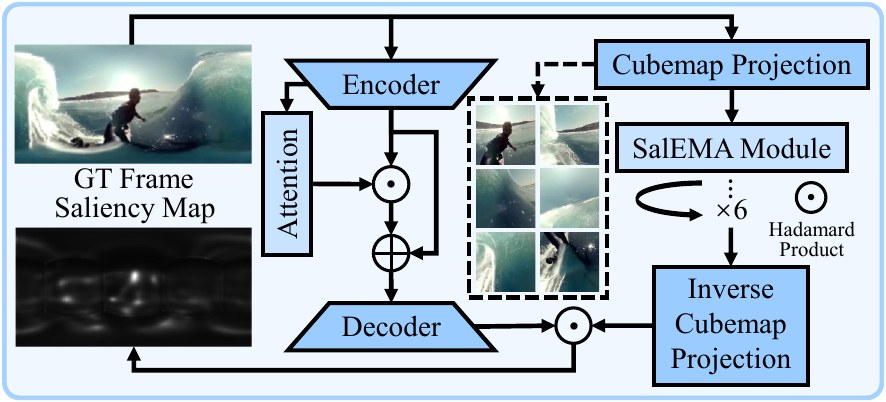}
\vspace{-13pt}
\caption{Saliency prediction module. The attention branch captures global information and the projection branch exploits local viewpoint details dynamically.}
\label{fig.7}
\vspace{-13pt}
\end{figure}

\subsection{Latitude-Saliency Adaptive Module (LSAM)}
Considering practical ODV watching habits, we introduce a latitude-saliency adaptive module (LSAM) to counteract latitude distortions while respecting viewing preferences. LSAM offline estimates latitude projection and saliency attention maps to assign weights in the loss function. Instead of fusing them as priors in the network, LSAM incurs zero additional cost at inference, making it deployment-friendly.

\subsubsection{Latitude-Related Projection Map}
As shown in Fig. \ref{fig.1}, ERP pixels are non-uniform. We quantify this distortion by the stretching ratio (STR) \cite{WS-PSNR}, which measures the local area variation after projection. From Appendix 1, the ERP STR is derived as:
\begin{equation}
\label{STR}
\text{STR}(u,v)=\cos(\phi)=\cos(v),
\end{equation}
Based on Eqs. \ref{STR}, we measure ERP distortions through latitude-related pixel weights ranging from 0 to 1:
\begin{equation}
\label{equ_4}
W_{lat}(u, v) = \cos ((v - (H / 2) + 0.5) \pi / H).
\end{equation}

\subsubsection{Saliency-Oriented Attention Map}
While ODVs span 360$^{\circ}$, human views fixate on limited regions. We therefore enhance high-frequency observed viewpoints to improve perception. Built upon SalEMA \cite{SalEMA, ATSal}, we utilize a two-branch network (Fig. \ref{fig.7}) to capture global context and regress viewpoint likelihood for predicting ODV saliency. Afterward, we multiply parallel results to get the saliency-oriented weight $W_{sal}(u, v)$.

\subsubsection{Latitude-Saliency Adaptive Loss Function}
As shown in Fig. \ref{fig_3}, LSAM is pre-computed to yield pixel-level weights. The composite loss steers the network toward attractive areas. Loss $L_{lat}$ and $L_{sal}$ are formulated as follows:
\begin{equation}
\label{equ_5}
L_{lat/Sal} =\mathbb{E}(W_{lat/Sal}\odot\left|I_{HR}-I_{SR}\right|),
\end{equation}
where $I_{HR}$ and $I_{SR}$ denote ground-truth HR and reconstructed SR patches. $\mathbb{E}$($\cdot$) is matrix averaging. The latitude-saliency adaptive loss function \emph{L} is defined as:
\begin{equation}
\label{equ_7}
L = \sqrt{{\|I_{HR}-I_{SR}\|}^2 + \varepsilon^2} + \alpha_2 L_{lat} + \beta_2 L_{sal},
\end{equation}
where the first item represents Charbonnier loss \cite{Charbonnier} and $\varepsilon$ is assigned as $10^{-3}$.

\begin{figure}[!t]
\centering
\includegraphics[width=0.98\columnwidth]{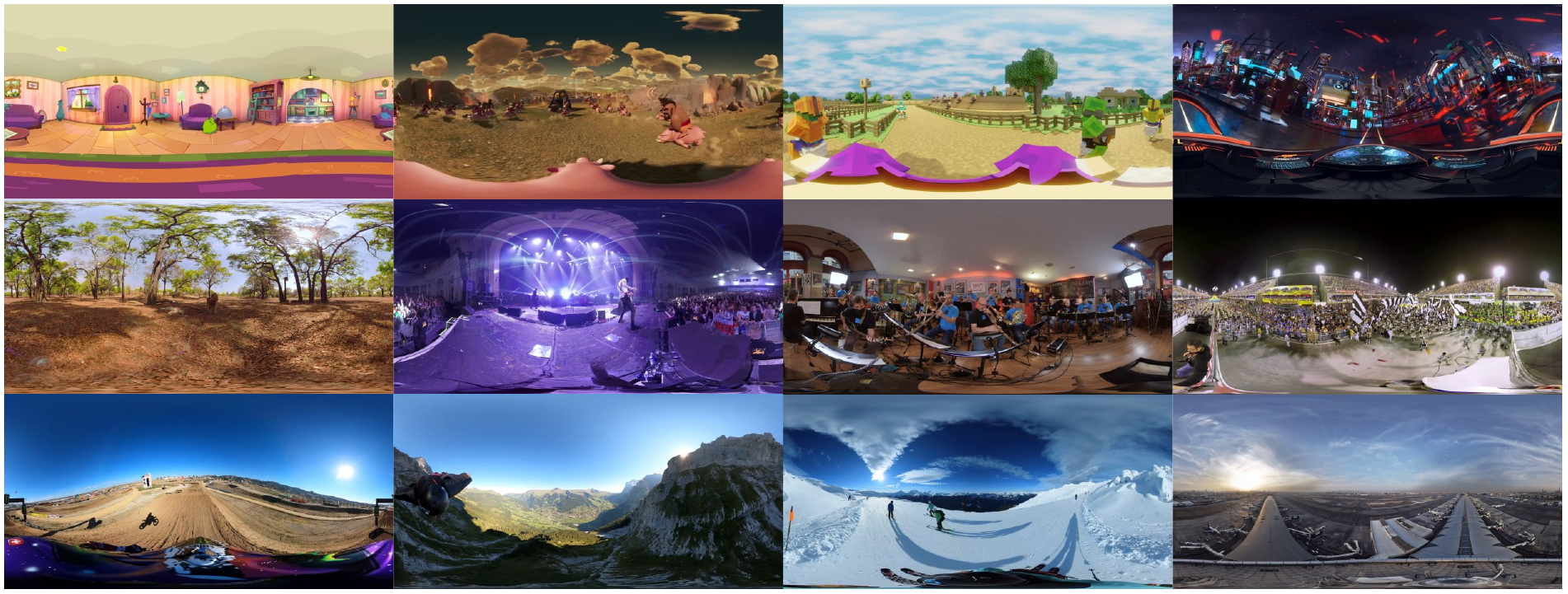}
\vspace{-6pt}
\caption{Examples of the proposed \emph{ODV-SR} dataset, including dynamic and static scenes.}
\label{fig_8}
\vspace{-14pt}
\end{figure}

\begin{figure*}[!t]
\centering
\includegraphics[width=2\columnwidth]{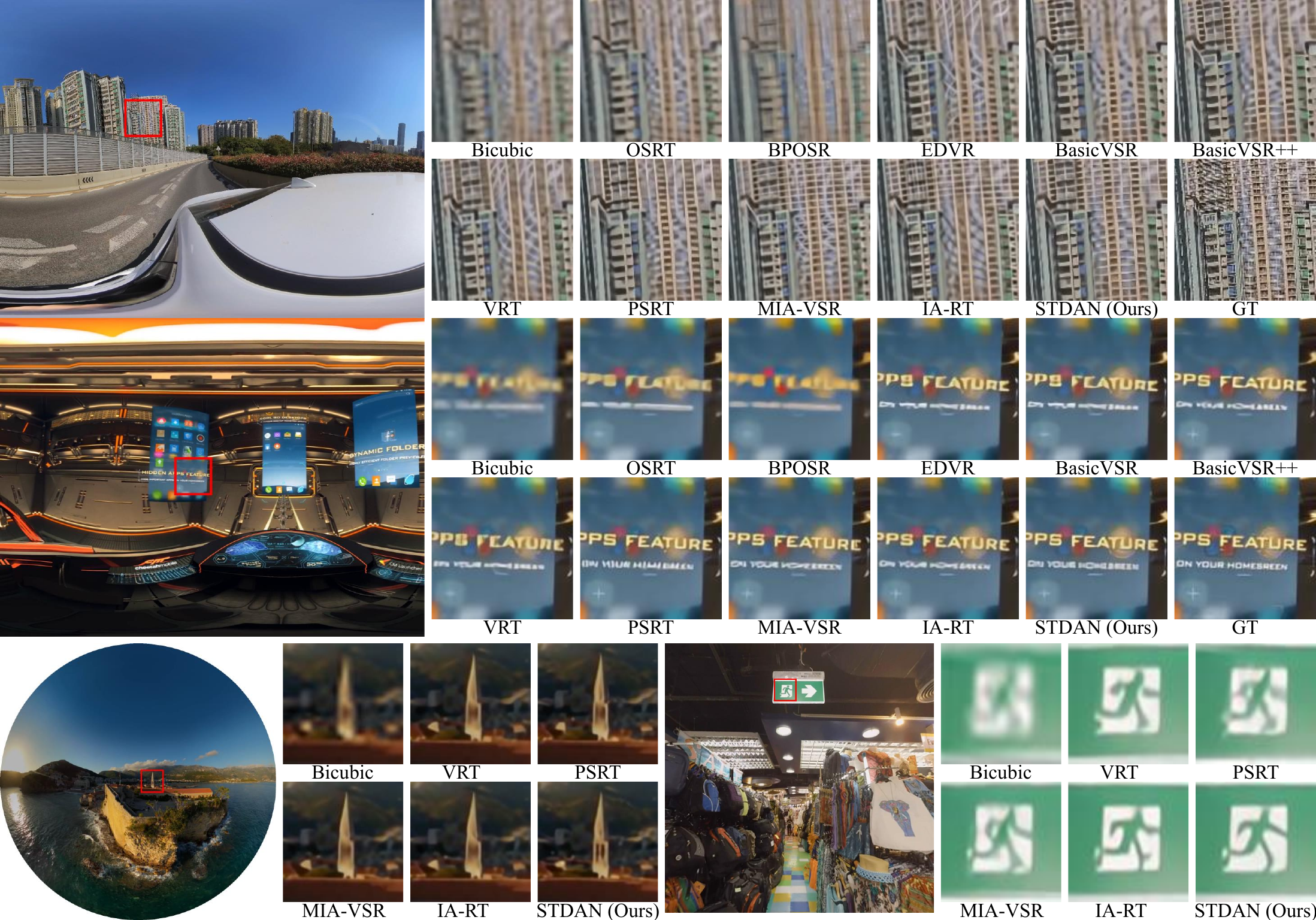}
\vspace{-4pt}
\caption{Qualitative $\times$4 ODV-SR comparison on ERP, fisheye, and perspective \emph{ODV-SR} dataset.}
\label{fig_9}
\vspace{-12pt}
\end{figure*}

\section{Experiments}

\subsection{Experimental Details}

Prior ODV datasets have few testing sequences (4, MiG \cite{SMFN}) or limited resolution (360 ${\times}$ 480, 360VDS\cite{S3PO}), which are unconvincing. Although NTIRE \cite{NTIRE} released a high-quality ODV360 dataset, its test set remains undisclosed. To facilitate training and testing, we gather a new ODV-SR dataset sourced and reorganized from public websites and the ODV360 dataset. The proposed \emph{ODV-SR} dataset comprises 270 train, 20 val, and 25 test clips at 1080 ${\times}$ 2160 resolution. As illustrated in Fig. \ref{fig_8}, the collection spans diverse scenes: landscapes, concerts, narration, sports, driving, movies, and virtual games, animation ODVs.

During training, HR frames are cropped to 256 ${\times}$ 256 patches and augmented with rotation and flip. Each clip contains 20 frames with a batch size of 2. STCA embeds 7 residual blocks. We train STDAN with Adam ($lr={10^{-4}}$) and CosineRestart for 300K iterations. The SpyNet is jointly fine-tuned at $\times$0.25 the STDAN learning rate. Training is conducted on 8 A800 GPUs in PyTorch.

\subsection{Comparison Results}
Interpolation method Bicubic, ODI-SR methods OSRT and BPOSR \cite{OSRT, BPOSR}, CNN-based VSR methods EDVR, BasicVSR, and BasicVSR++ \cite{EDVR, BasicVSR, BasicVSR++}, Transformer-based VSR method, VRT, PSRT, IA-RT, and MIA-VSR, \cite{VRT, Patch, IART, MIAVSR}, ODV-SR methods SMFN and S3PO \cite{SMFN, S3PO}, and our STDAN are comprehensively evaluated to compare ODV-SR performance. PSNR, SSIM, and ERP-specific WS-PSNR \cite{WS-PSNR}, WS-SSIM \cite{WS-SSIM} are calculated on the luma (Y) channel. We fairly retrain reproducible video models on the ODV360 dataset, otherwise, results are migrated from the 360VDS dataset and MiG dataset as reported in papers.
% Because Transformer-based models have high computational complexity, we keep other training configurations the same and maximize the sequence length on GPUs to simulate practical application scenarios. 

%\begin{table}[!t]
%\caption{Computational efficiency comparison of $\times$4 ODV-SR. Params denotes the number of parameters.}
%\vspace{-4pt}
%\label{tab_efficiency}
%\centering
%\footnotesize
%\renewcommand\arraystretch{0.85}{
%\setlength{\tabcolsep}{3pt}
%\begin{tabular}{c|c|c|c|c}
%\hline
%\multirow{2}{*}{Method}&360-SS\cite{360-SS}&OSRT\cite{OSRT}&EDVR\cite{EDVR}&BasicVSR\cite{BasicVSR}\\
%&BasicVSR++\cite{BasicVSR++}&VRT\cite{VRT}&PSRT\cite{Patch}&STDAN(ours)\\
%\hline
%\multirow{2}{*}{Params(M)}&0.7&1.1&20.0&6.3\\
%&7.0&29.1&12.3&27.8\\
%\hline
%\multirow{2}{*}{Runtime(s)}&0.116&0.833&0.353&0.155\\
%&0.202&4.199&5.335&0.696\\
%\hline
%\end{tabular}}
%\vspace{-8pt}
%\end{table}

\subsubsection{Computational Efficiency Comparison}
Computational efficiency on the \emph{ODV-SR} dataset is summarized in Fig. \ref{fig.2}. Relative to Transformer-based VSR approaches, VRT, PSRT, MIA-VSR, and IA-RT, STDAN achieves a faster running speed while preserving the best SR quality, striking an effective complexity-performance trade-off. Detailed parameters, FLOPs, and runtime are reported in Appendix 2.1.

\subsubsection{Quantitative Comparison}
As displayed in Tab. \ref{tab_ODV_SR}, our STDAN achieves the best or second-best quantitative metrics on all benchmarks, demonstrating its superior capacity. Notably, the HR-grade \emph{ODV-SR} dataset offers results that are closer to real-world conditions. While Transformer-based MIA-VSR and IA-RT yield competitive performance, they incur markedly higher computational overhead.
% Tab. \ref{tab_360VDS} displays the $\times$4 SR experimental results on the 360VDS dataset and MiG dataset. STDAN also achieves the best performance. These experimental results demonstrate the superior efficiency of STDAN over other SOTA methods, especially on WS-PSNR. It is worth noting that the Transformer-based VRT yields the second-best results, but it spends six times as much inference time as STDAN.

%\begin{figure}[!t]
%\centering
%\includegraphics[width=\columnwidth]{Image/10.pdf}
%\vspace{-16pt}
%\caption{Qualitative $\times$4 ODV-SR comparison on fisheye (the first image) and perspective (the second image) \emph{ODV-SR} dataset.}
%\label{fig_10}
%\vspace{-16pt}
%\end{figure}

\subsubsection{Qualitative Comparison}
Fig. \ref{fig_9} presents visual results from representative VSR methods. Upon inspection of zoomed-in regions, only STDAN restores grids and stripes of buildings clearly. In the second scene, the small white letters ``YOUR'' become most legible in the STDAN reconstruction. Beyond the ERP format, we further visualize results on fisheye and perspective ODVs. On such formats, STDAN continues to rebuild visually more pleasing textures. These observations corroborate that STDAN reconstructs finer structural details and surpasses current methods. More comparison results can be found in Appendix 2.2.

\subsubsection{Spatio-Temporal Consistency Comparison}
\begin{figure}[!t]
\centering
\includegraphics[width=\columnwidth]{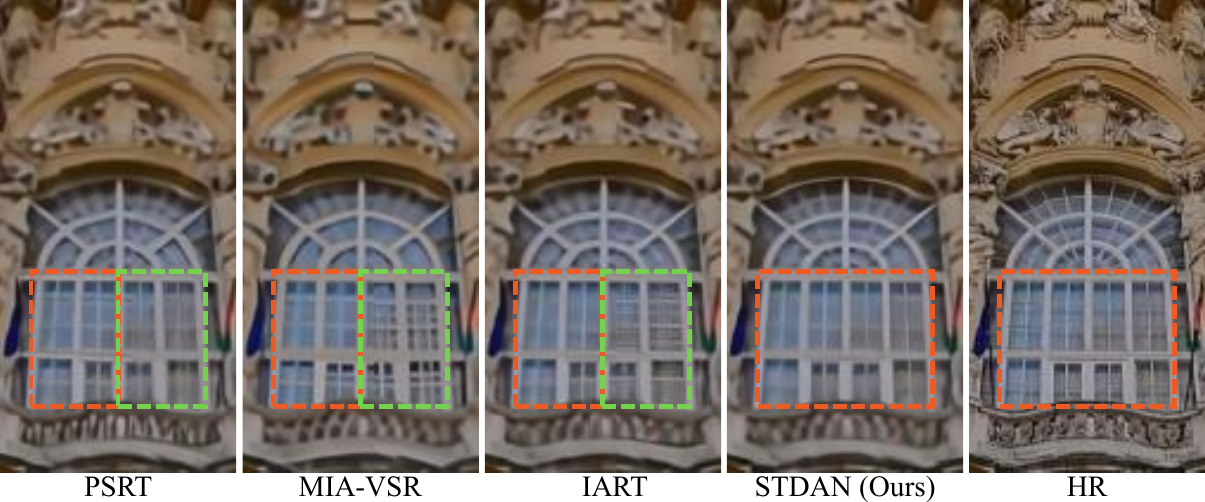}
\vspace{-16pt}
\caption{Qualitative $\times$4 ODV-SR comparison on ERP boundaries in \emph{ODV-SR} dataset.}
\label{fig_11}
\vspace{-6pt}
\end{figure}

As discussed in Sec. STCA, we introduce spatio-temporal joint alignment to modulate discontinuities flexibly. First, we evaluate spatial consistency across ERP boundaries. According to Fig. \ref{fig_5} (a) and \ref{fig_11}, we stich the left and right margins into a unified perspective view. Within the highlighted rectangular regions, it is clearly evident that STDAN maintains identical window structures on both sides of the seam, whereas competing approaches reconstruct dissimilar neighboring windows with obvious visual discontinuities. 

The warping-error metric $\text{E}_{warp}^*$ \cite{WE} quantifies temporal consistency. As illustrated in Tab. \ref{tab_ODV_SR}, STDAN attains the lowest $\text{E}_{warp}^*$, evidencing its generation of smooth and flicker-free ODVs. Moreover, VMAF \cite{VMAF} is a widely adopted video quality assessment that jointly measures spatial fidelity and temporal motion coherence. We exploit it to examine the spatio-temporal performance of ODV-SR. From Tab. \ref{tab_ODV_SR}, STDAN receives the highest VMAF score. To intuitively indicate the temporal stability improvement of STDAN, we include temporal profiles of different algorithms in Appendix 2.3. We also conduct a user study in Appendix 2.4.% The temporal profiles are obtained by stitching horizontal rows at the same location in consistent frames. It is observed that STDAN achieves more seamless performance. We also present multi-frame reconstruction comparison in Appendix.

%\begin{figure}[!t]
%\centering
%\includegraphics[width=\columnwidth]{Image/12.pdf}
%\vspace{-16pt}
%\caption{Temporal profile comparison of $\times$4 ODV-SR on ERP \emph{ODV-SR} dataset.}
%\label{fig_12}
%\vspace{-12pt}
%\end{figure}

\begin{figure}[!t]
\centering
\includegraphics[width=\columnwidth]{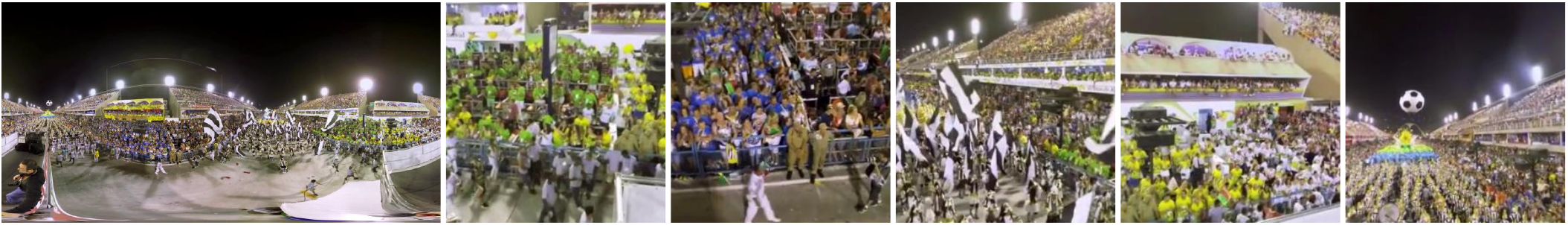}
\vspace{-16pt}
\caption{Top-5 viewpoints of an ERP ODV. The observed attractive viewpoints are projected from ERP to perspective.}
\label{fig_13}
\vspace{-12pt}
\end{figure}

\subsubsection{Viewpoint Comparison}
When watching 360$^{\circ}$ ODVs, viewers naturally focus on appealing local regions. To assess practical perceptual quality, we sample the Top-5 high-frequency observed viewpoints with a SURF-based detector \cite{Viewpoint} and calculate their PSNR and SSIM. An illustrative example is provided in Fig. \ref{fig_13}. By achieving 0.14 dB PSNR gains over IA-RT in Tab. \ref{tab_ODV_SR}, STDAN performs better on salient viewpoints where it matters most to human observers.

\section{Ablation Study}
In this section, we perform ablation studies to validate each proposed module, comparing the $\times$4 SR baseline against variants that incrementally include individual components.

\begin{table}[!t]
\footnotesize
\centering
\renewcommand\arraystretch{0.89}{
\setlength{\tabcolsep}{1.2pt}
\begin{tabular}{c|c|c}
\hline
\multicolumn{2}{c|}{\multirow{2}{*}{\centering Models with Proposed Modules}}&\emph{ODV-SR} Dataset\\
\multicolumn{2}{c|}{}&WS-PSNR / WS-SSIM\\
\hline
0&Baseline&29.83 / 0.8621\\
1&+Latitude Distortion Map&29.93 / 0.8643\\
2&+Saliency Attention Map&29.91 / 0.8641\\
3&+Spatio Distortion Alignment&29.85 / 0.8635\\
4&+Cylinder Position Encoding& 29.88 / 0.8639\\
5&+Interlaced Multi-frame Reconstruction&30.06 / 0.8665\\
6&+Interlaced Multi-frame Fusion&30.02 / 0.8667\\
7&+Data Augmentation&30.04 / 0.8653\\
\hline
\end{tabular}}
\vspace{-5pt}
\caption{$\times$4 ODV-SR ablation study on different modules.}\label{tab_4}
\vspace{-5pt}
\end{table}

\subsection{Effects of Proposed Modules}
Tab. \ref{tab_4} summarizes the ablation results. Model 0 is the baseline and Models 1-7 add one module at a time. Specifically, Models 1 and 2 with the refined loss bring about 0.1 dB WS-PSNR gains. Model 3 exceeds the baseline by 0.13 dB, and its optimized variant Model 4 gains another 0.03 dB. Models 5 and 6 achieve about 0.2 dB improvements with iteration in three temporal sequences. A similar conclusion can be observed in WS-SSIM. These results indicate the effectiveness of the proposed modules, which can enhance ODV-SR performance. Besides, Model 7 boots WS-PSNR by 0.21 dB, confirming the value of the proposed \emph{ODV-SR} dataset. Moreover, we visualize $\times$4 SR results of Models 1-7 in Appendix 2.5 to provide an intuitive comparison.

%Furthermore, we visualize $\times$4 SR results of Models 0 to 6. As presented in Fig. \ref{fig_17}, models with proposed modules reconstruct sharper frames, especially Models 2, 4, and 6 successfully recover the letters ``GHT'' which are blurred in the baseline. Additionally, Models 3 and 5 restore clear letters ``GHLI''. Compared with the baseline, the proposed modules can produce more accurate texture details with a pleasing visual feeling. In terms of patch PSNR and SSIM performance, Model 1 degrades since it inclines to emphasize equator areas. Other models all make progress and Model 5 has an obvious better SSIM, which immediately improves the visual quality.

\subsection{Effects of Different Fusion and Weight Parameters}
As discussed in IMFR and LSAM Section, we preset fusion parameters $\alpha_1, \beta_1 = 0.01$ and balance latitude and saliency weights with $\alpha_2, \beta_2 = 0.1$. As shown in Tab. \ref{tab_5}, these default values are from a grid search and yield the best performance.

%\begin{figure}[!t]
%\centering
%\includegraphics[width=1.0\columnwidth]{ICCV2025-Author-Kit-Feb/Image/13.pdf}
%\caption{Subjective ablation study of $\times$4 ODV-SR with different modules on ODV-SR dataset. The PSNR and SSIM of the local regions are provided for a more intuitive comparison.}
%\label{fig_17}
%\vspace{-8pt}
%\end{figure}

\begin{table}[!t]
\footnotesize
\centering
\renewcommand\arraystretch{0.89}{
\begin{tabular}{c|c|c|c}
\hline
\multicolumn{2}{c|}{Fusion and Weight}&\multicolumn{2}{c}{\emph{ODV-SR Dataset}}\\ 
\cline{3-4}
\multicolumn{2}{c|}{Parameters}&\multicolumn{2}{c}{WS-PSNR(dB) / WS-SSIM}\\
\hline
\multicolumn{2}{c|}{$\alpha_1$, $\beta_1$ = 0}&\multicolumn{2}{c}{29.83 / 0.8621}\\
\multicolumn{2}{c|}{$\alpha_1$, $\beta_1$ = 0.001}&\multicolumn{2}{c}{\underline{29.98} / 0.8657}\\
\multicolumn{2}{c|}{$\alpha_1$, $\beta_1$ = 0.01}&\multicolumn{2}{c}{\textbf{30.02} / \underline{0.8667}}\\
\multicolumn{2}{c|}{$\alpha_1$, $\beta_1$ = 0.1}&\multicolumn{2}{c}{29.69 / \textbf{0.8669}}\\
\hline
$\alpha_2$ = 0&$\beta_2$ = 0&29.83 / 0.8621&29.83 / 0.8621\\
$\alpha_2$ = 0.01&$\beta_2$ = 0.01&\underline{29.90} / \textbf{0.8644}&\underline{29.86} / \underline{0.8637}\\
$\alpha_2$ = 0.1&$\beta_2$ = 0.1&\textbf{29.93} / \underline{0.8643}&\textbf{29.91} / \textbf{0.8641}\\
$\alpha_2$ = 1&$\beta_2$ = 1&29.81 / 0.8610&29.83 / 0.8615\\
\hline
\end{tabular}}
\vspace{-4pt}
\caption{$\times$4 ODV-SR comparison of different fusion and weight parameters.}\label{tab_5}
\vspace{-13pt}
\end{table}

\section{Conclusion}

In this paper, we proposed a Spatio-Temporal Distortion Aware Network (STDAN) for omnidirectional video super-resolution (ODV-SR) in practical scenarios. First, we designed a spatio-temporal continuous alignment module with omni-positional coding to offset spatio-temporal distortions and discontinuities. Next, we presented an interlaced multi-frame reconstruction module to enhance the temporal consistency of restored ODVs. Finally, we introduced a latitude-saliency adaptive module to weigh regions with visually sensitive texture and human-watching interest. We further collected a diverse \emph{ODV-SR} dataset and developed comprehensive experiments on it. Extensive experimental results on different datasets demonstrated that our STDAN achieves the best ODV-SR performance and is application-friendly with faster speed and real-world viewpoint enhancement.

\bibliography{aaai2026}

\cleardoublepage

\includepdf[pages=-]{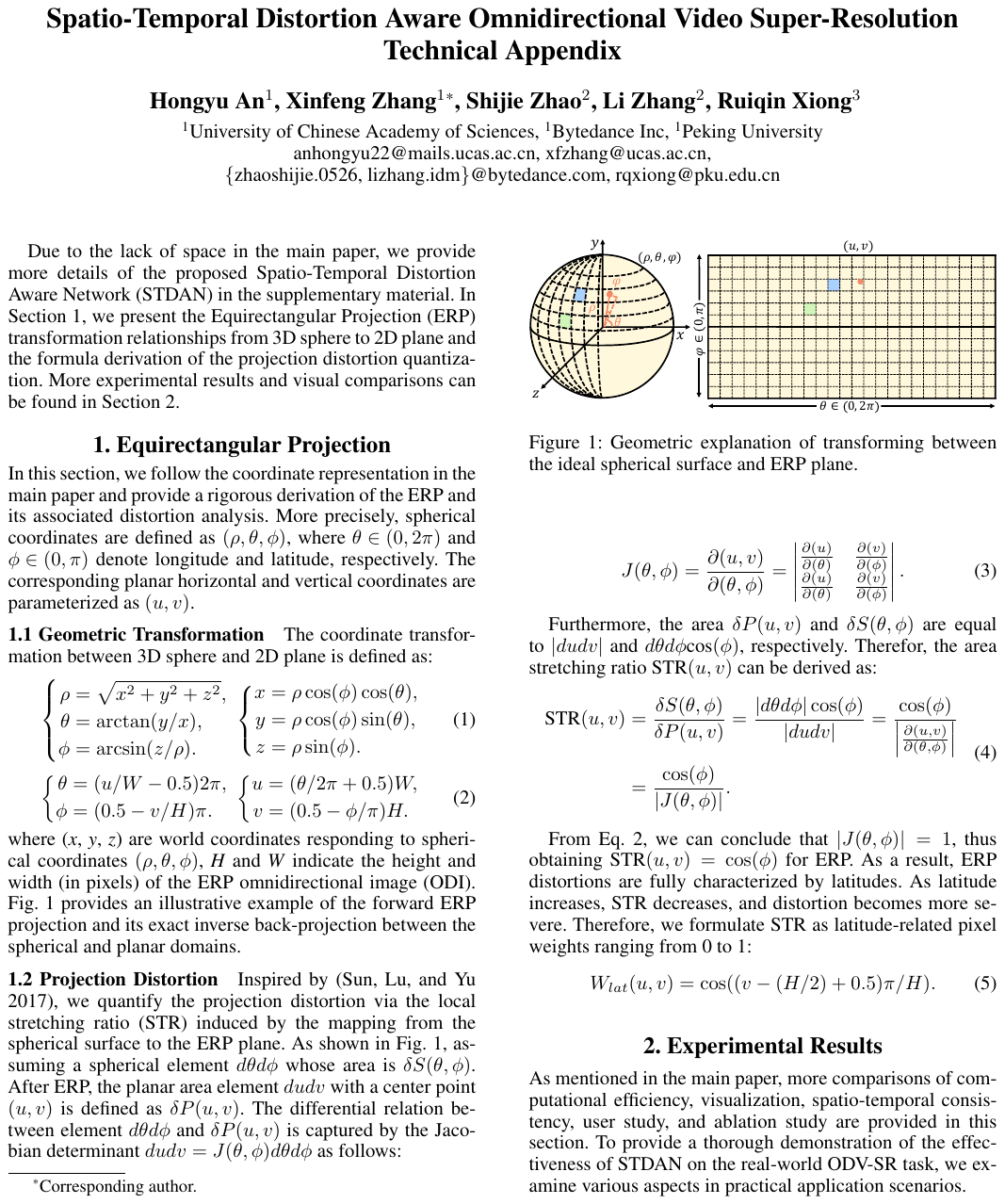}

\end{document}